\documentclass{article}

\usepackage{microtype}
\usepackage{graphicx}
\usepackage{subfigure}
\usepackage{booktabs} 

\usepackage{hyperref}

\usepackage[accepted]{latex/icml2023}

\usepackage{amsmath}
\usepackage{amssymb}
\usepackage{mathtools}
\usepackage{amsthm}

\usepackage[capitalize,noabbrev]{cleveref}

\theoremstyle{plain}

\theoremstyle{definition}

\theoremstyle{remark}

\usepackage[textsize=tiny]{todonotes}

\icmltitlerunning{Towards A Unified Neural Architecture for Visual Recognition and Reasoning}

\begin{document}

\twocolumn[
\icmltitle{Towards A Unified Neural Architecture for Visual Recognition and Reasoning}



\icmlsetsymbol{equal}{*}

\begin{icmlauthorlist}
\icmlauthor{Calvin Luo}{brown}
\icmlauthor{Boqing Gong}{google}
\icmlauthor{Ting Chen}{google}
\icmlauthor{Chen Sun}{brown}
\end{icmlauthorlist}

\icmlaffiliation{brown}{Brown University}
\icmlaffiliation{google}{Google. Work performed when Calvin Luo interned at Google Research}

\icmlcorrespondingauthor{Calvin Luo}{calvin\_luo@brown.edu}
\icmlcorrespondingauthor{Chen Sun}{chensun@brown.edu}

\icmlkeywords{visual reasoning}

\vskip 0.3in
]



\printAffiliationsAndNotice{}  

\begin{abstract}
Recognition and reasoning are two pillars of visual understanding.  However, these tasks have an imbalance in focus; whereas recent advances in neural networks have shown strong empirical performance in visual recognition, there has been comparably much less success in solving visual reasoning.  Intuitively, unifying these two tasks under a singular framework is desirable, as they are mutually dependent and beneficial.  Motivated by the recent success of multi-task transformers for visual recognition and language understanding, we propose a unified neural architecture for visual recognition and reasoning with a generic interface (e.g., tokens) for both.  Our framework enables the principled investigation of how different visual recognition tasks, datasets, and inductive biases can help enable spatiotemporal reasoning capabilities.  Noticeably, we find that object detection, which requires spatial localization of individual objects, is the most beneficial recognition task for reasoning.  We further demonstrate via probing that implicit object-centric representations emerge automatically inside our framework. Intriguingly, we discover that 
certain architectural choices such as the backbone model of the visual encoder have a significant impact on visual reasoning, but little on object detection.  Given the results of our experiments, we believe that visual reasoning should be considered as a first-class citizen alongside visual recognition, as they are strongly correlated but benefit from potentially different design choices.
\end{abstract}
\section{Introduction}

Modern advances in convolutional neural networks~\citep{lecun1995convolutional} have demonstrated great proficiency in visual recognition tasks, such as image classification~\citep{krizhevsky2012imagenet}, object detection~\citep{ren2015faster}, and instance segmentation~\citep{he2017mask}. More recently, the Transformer architecture~\citep{vaswani2017attention}, initially designed for language understanding tasks, has demonstrated competitive performance on image~\citep{dosovitskiy2020image} and video~\citep{arnab2021vivit} recognition tasks. Although the performance gains of the Transformer over alternative architectures for visual perception is still under debate~\citep{mlpmixer, convnext}, one concrete benefit of the Transformer is its versatility in modeling diverse input modalities~\citep{jaegle2021perceiver,alayrac2022flamingo}, and its flexibility in unifying a wide range of perception and reasoning tasks as sequence prediction~\citep{raffel2019exploring,chen2021pix2seq}.
Indeed, scaling Transformers massively in size, compute, and data has enabled many exciting recent developments such as multi-modal ``foundation models''~\citep{foundation_models} and multi-task ``generalist agents''~\citep{reed2022generalist}.
Motivated by their success, our paper investigates how this paradigm, where a neural network is employed agnostic of task-specific inductive biases, can be used to build a unified model that can perform both visual recognition and visual reasoning tasks.

Visual perception and reasoning are the two pillars of visual understanding. Despite the rapid progress on visual question answering~\citep{antol2015vqa,hudson2019gqa}, primarily due to Transformer networks and large-scale pre-training~\citep{alayrac2022flamingo}, it has been observed that state-of-the-art neural networks designed for visual recognition have much less success in solving even the most basic reasoning tasks that require causal inference~\citep{yi2019clevrer,zhang2021acre} or the notion of object permanence~\citep{girdhar2019cater}. 
Our study focuses on these visual reasoning tasks, which require understanding object attributes, relationships, and dynamics. Specifically, we use two diagnostic benchmarks: CATER~\citep{girdhar2019cater}, where a model must learn object permanence to track occluded objects in videos, and ACRE~\citep{zhang2021acre}, which requires a model to perform causal inference in a setup inspired by the famous Blicket experiment~\citep{gopnik2000detecting}.


Researchers~\citep{greff2020binding, santoro2021symbolic} have advocated that an object-centric, symbol-like representation is essential for compositional generalization required by reasoning tasks. It was also commonly believed that although neural networks may excel at extracting objects and their attributes given a pre-defined vocabulary, the reasoning module should be based on symbolic approaches~\citep{mao2019neuro}. However, the seminal work ALOE by~\citet{ding2021attention} demonstrated that a Transformer neural network can not only perform reasoning, but also even significantly outperform its neuro-symbolic counterparts at times. Similar to neuro-symbolic approaches, ALOE still requires pre-computed object-centric representations~\citep{burgess2019monet}, where visual recognition is a separate component and arguably requires task-specific knowledge to design.

In this work, we propose a unified framework for recognition and reasoning, where visual representations can be extracted, organized, and routed dynamically to solve both sets of tasks in parallel.
We are particularly interested in understanding whether visual representations are organized in an \textit{object}-centric fashion in the intermediate layers, which conceptually would facilitate compositional generalization. We follow the notions by~\citet{greff2020binding} and refer to symbol-like entities as objects. In the context of our work, they can correspond to objects, object parts, or spatiotemporal object trajectories.  We hypothesize that the choice of recognition task can help promote the emergence of such implicit object-centric representations; we further conjecture that tasks involving spatiotemporal understanding such as object detection are particularly helpful.

Under our proposed unified framework, we are able to achieve competitive performance on CATER for object tracking with occlusion, and reasonable performance on ACRE for causal inference.  Furthermore, our framework enables the principled comparison between the effects of different design decisions on reasoning capabilities.  When ablating over different visual recognition tasks, we find that the object detection task is critical for better reasoning capability of the model; we also qualitatively visualize via probing that object-centric representations emerge in the middle of the network.  We believe these results show encouraging signals towards building a unified and generic multi-task framework for recognition and reasoning.  Our ablation study also reveals a surprising observation that while the object detection performance is robust across different choices of neural architectures, such as the use of ViT~\citep{dosovitskiy2020image} or a ResNet~\citep{he2016deep} visual encoder, different inductive biases can lead to significant performance gaps on the model's reasoning performance.  As most of the existing visual multi-task models~\citep{alayrac2022flamingo} or benchmarks~\citep{vtab} focus on visual recognition, we hope our findings can bring awareness to treating visual reasoning as a first-class citizen when designing neural architectures and benchmarking model performances.


Overall, our main contributions are as follows:
\begin{itemize}
    \item We propose a unified end-to-end architecture that performs reasoning alongside object detection, eliminating the need for pipeline-style approaches.
    \item We utilize our architecture to investigate different considerations, such as inductive biases or visual recognition objectives, when designing solutions for spatiotemporal reasoning.
    \item We observe that the object detection task leads to the emergence of object-centric representation, which we hypothesize improves reasoning.  We visualize and investigate these representations through probing.
\end{itemize}

\begin{figure*}[t]
  \centering
  \includegraphics[width=.95\linewidth]{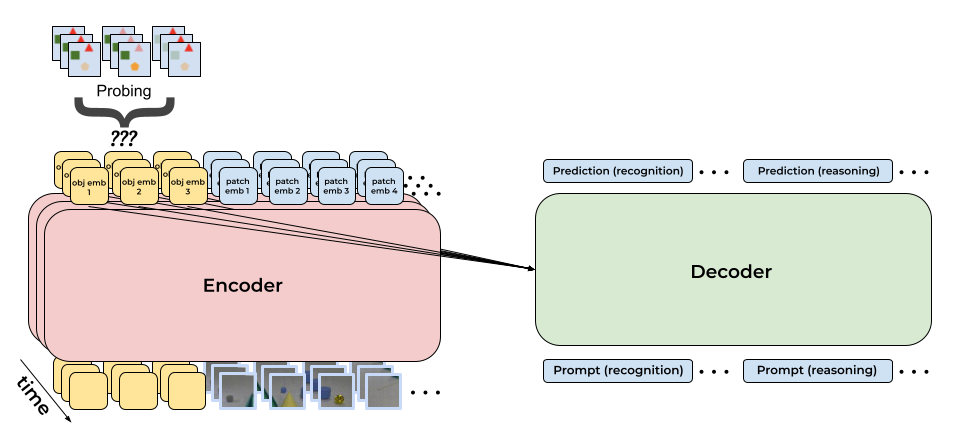}
  \caption{An illustration of our proposed unified neural architecture for visual recognition and reasoning. Each image or its encoded feature map is first broken into patches and fed to a Transformer encoder. We employ a few ``slot'' tokens (yellow) which condense the visual information and pass them to the decoder. The slot tokens from different frames of a video are concatenated over time. The Transformer decoder then autoregressively predicts task-specific output sequences according to their corresponding ``prompts'', such as object labels, (quantized) object locations, or answers to a reasoning task. To understand if the slot tokens encode object-centric representations we train a probing classifier, also modeled as an autoregressive decoder under our framework, and ask it to predict object locations given a randomly selected slot token. An object-centric token is expected to successfully detect only one or few objects.}
  \vspace{-1em}
  \label{fig:architecture}
\end{figure*}

\section{Related Work}

\textbf{Visual reasoning.} The huge success of deep learning for image understanding tasks have prompted researchers to tackle visual reasoning tasks, which are more challenging, in the form of visual question answering~\citep{antol2015vqa,lu2016hierarchical,hudson2019gqa}, visual commonsense extraction~\citep{yatskar2016stating} and reasoning~\citep{visual_commonsense}, or visual dialog~\citep{das2017visual}. While neural networks have achieved remarkable success in these benchmarks that require reasoning, it has also been observed that dataset bias and language bias~\citep{goyal2017making} make it harder to concretely measure progress. This has lead to a suite of synthetic, diagnostic datasets~\citep{johnson2017clevr,yi2019clevrer,girdhar2019cater,zhang2021acre} to benchmark visual reasoning. These benchmarks highlighted the limitations of end-to-end trained neural networks, and often advocated for neural-symbolic methods~\citep{mao2019neuro,hudson2019learning}. However, such neuro-symbolic approaches often require task-specific knowledge to define the ``symbols'' and the ``programs'', making them more challenging to generalize across tasks. Recently,~\citet{ding2021attention} demonstrated that an end-to-end trained Transformer network can indeed perform reasoning, given that its inputs are ``symbol-like'' object segments. Our paper aims to take one step further and investigate if it is possible to build a unified framework for visual recognition and reasoning without committing to a task-specific inductive bias at the model's inputs, which would make the framework more general.

\textbf{Object-centric representation} is an essential component for both neuro-symbolic~\citep{mao2019neuro} and Transformer-based~\citep{ding2021attention} frameworks for visual reasoning. \cite{greff2020binding} argue that a reasoning framework consists of three stages, namely \textit{segregation}, \textit{representation}, and \textit{composition}. The \textit{segregation} stage is responsible for extracting objects (or other symbol-like entities), and can be implemented with a supervised object detector~\citep{mao2019neuro}, such as Mask R-CNN~\citep{he2017mask}. There is also a line of research to learn low-level features correlated with objects, such as textures~\citep{geirhos2018imagenet, hermann2020origins, olah2017feature}, or even object detectors~\citep{burgess2019monet,locatello2020object,caron2021emerging} with self-supervised objectives. Researchers working on model interpretability also attempt to visualize the internal activations of a trained neural network, and inspect if they correspond to objects or object parts~\citep{bau2017network,bau2018gan}. Our work takes a supervised framework~\citep{chen2021pix2seq} for its general-purpose transformer-based architecture, which can be easily generalized to perform reasoning.

\textbf{Multi-task Transformers} have achieved tremendous success~\citep{alayrac2022flamingo,reed2022generalist,foundation_models} on tasks that require visual perception and control. Particularly appealing is the their versatility in incorporating multi-modal inputs with minimal modality-specific assumptions~\citep{akbari2021vatt,jaegle2021perceiver}, and its flexibility to express the training objectives of diverse tasks all as sequence prediction~\citep{raffel2019exploring,chen2021pix2seq}. Large-scale pre-training, both for the number of model parameters, and for the amount of (unlabeled) training data, is also found to be crucial~\citep{foundation_models}. Our generalized framework, which unifies the object detection and visual reasoning tasks, belongs to the model family of multi-task Transformers.

\section{Method}
\label{method}

We seek to unify both recognition and reasoning tasks under a common framework; intuitively, we can do so by formulating everything as a sequence-to-sequence prediction task. The input sequence can be image patches and optionally a question that requires reasoning to solve, and the output sequence can be parameterized for recognition (e.g. as proposed by~\citet{chen2021pix2seq}), or reasoning (e.g. an index for multiple-choice questions, or a plain text answer). We can then use an encoder-decoder framework to model both recognition and reasoning tasks together.



\textbf{Architecture.} Our encoder-decoder framework is implemented with the Transformer networks~\cite{vaswani2017attention}. The encoder can be a Vision Transformer which takes \textit{patchified} images as inputs. The input patch embeddings can be implemented as linear projections~\citep{dosovitskiy2020image}, or as the intermediate features of a ResNet~\citep{he2016deep}.  The autoregressive decoder conditions on the output embeddings of the encoder, and applies minimal task-specific priors. As a result, the proposed framework is general, as it can can not only handle recognition tasks, but also reasoning tasks over time in a unified fashion.
In order to perform spatio-temporal reasoning, we enable our framework to handle the video domain by simply \textit{stacking} the encoder structure over time; we reuse the same weights, but handle each frame independently. To encourage the emergence of object-centric in the encoder network, we utilize a cross-attention bottleneck to force the representation of the input to be encoded in the form of a few compact tokens. Denoted by the yellow tokens in Figure \ref{fig:architecture}, only these tokens are passed forward to the decoder for conditioning.  We interpret these bottleneck tokens as slots, and hypothesize that they bind to object-centric information.  We subsequently verify this explicitly via probing experiments in Section~\ref{probing}. The slot tokens are extracted for each video frame independently. They are then concatenated together, preserving temporal order, and positional encodings are added before they are passed to the decoder for conditioning.

The encoder-decoder framework is optimized with the sequence completion objective. For visual reasoning, the output sequence corresponds to plain-text answers or indices when the candidate answers are provided. For visual recognition, the output sequences may correspond to image labels or object bounding boxes. For example, Pix2Seq~\citep{chen2021pix2seq} is an example of our proposed framework where object detection is the sole task to perform.  In its default formulation, Pix2Seq processes a single image as a sequence-to-sequence task; visual patches are fed in as inputs to an encoder, and bounding box predictions are generated as an output sequence via an autoregressive decoder. The Pix2Seq objective for object detection is further generalized to the panoptic segmentation task by \citet{chen2022generalist}.



\textbf{Multi-task training.} We now extend our previously described architecture to handle multiple tasks simultaneously.  There are two approaches that can accomplish this: joint training, and alternating optimization. 
For joint training, recall that the decoder portion of the encoder-decoder framework conditions on the output of the encoder, as well as a task-specific prompt.  By feeding in multiple prompts over one step, the decoder can then jointly output the predictions for multiple tasks; the loss can then be computed to optimize the model with respect to several objectives at once.  In the alternating optimization approach, our proposed framework iteratively switches between optimizing for different tasks at fixed interval steps.  Intuitively, since an ordering to tasks might potentially be useful, this may result in better performing unified models.  For example, if we expect that the representations learned by the perceptual tasks might be subsequently useful for visual reasoning, we might prefer to first train on the perceptual task for a number of iterations before alternating.  Switching back from optimizing for the visual reasoning task to the perceptual task can also be useful; it allows for the implicit representations to be supervised with reasoning-specific signals.  In the extreme case, where we only perform one switch between tasks, this is equivalent to pre-training our model on a perceptual task and finetuning it on a reasoning task. We explore different multi-task training strategies in our experiments and find that for us, a single switch from object detection to reasoning yields the best performance on the reasoning tasks.

\textbf{Representation probing.} We are interested in investigating the encoder's slot embeddings to determine if they exhibit some object-centric or localized properties; as the decoder directly references them to solve tasks, their representational quality directly impacts performance.  One way to evaluate the information contained within the embeddings is through probing; after designating a set of embeddings of interest (or combinations thereof, in the general case), we can probe if they contain object-centric information such as bounding box coordinates or class labels.  The object detection training procedure can be reused under our framework with two changes: we randomly sample embeddings from the investigation set at each training iteration to attempt the object detection task, and we freeze the encoder so that the representations are purely evaluated and are not modified or supervised further.  After the decoder is trained to perform probing in such a fashion, we can apply it during inference to evaluate and visualize every embedding of interest.
Intuitively, if the encoder chooses to organize visual information in an object-centric fashion, it is expected that every probed embedding can successfully detect only one or few objects in an image, as information about other objects and their corresponding attributes are \textit{routed} to other embeddings. 

We note that since our framework also models probing as sequence prediction, it can be easily extended to incorporate arbitrary ground-truth probing information available in a given domain, such as color and material, through constructing an appropriate target prediction sequence.  Overall, our  framework naturally enables flexible probing functionality, thus enabling users to interpret and analyze the internal representations of the model with minimal adjustment.

\section{Experiments}

We consider two spatiotemporal reasoning tasks, CATER~\citep{girdhar2019cater} and ACRE~\citep{zhang2021acre}, and perform ablations over them.  The encoder and decoder of the model is also initialized from a pre-trained ImageNet classification checkpoint, on top of which all visual and reasoning tasks are further optimized upon.
All experiments were performed on 32 TPUs per run.


\subsection{Recognition Tasks Setup}
We begin by optimizing visual recognition tasks under our framework, in a supervised fashion.  There are a variety of visual recognition tasks of interest, ranging from classification to spatial localization objectives such as object detection.  Furthermore, the visual dataset associated with such tasks is also ablated over.  A thorough comparison study between these choices, and how they affect higher-level reasoning capabilities, are explored in Section~\ref{sec:pretraining}.

We observe that training on specific recognition tasks requires only a few thousand steps, as the encoder and decoder of the model are already initialized from a pre-trained ImageNet classification checkpoint.

\subsection{CATER Reasoning Setup}

\begin{figure}
  \centering
  \includegraphics[width=0.99\linewidth]{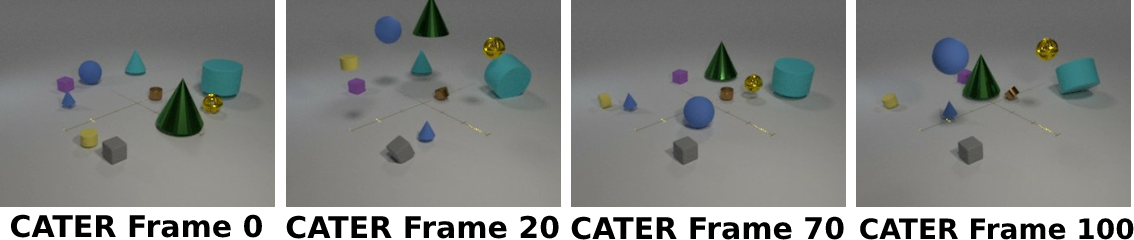}
  \caption{Example frames from a CATER video. 
  The task is to determine the final location of the golden snitch, which may be occluded or covered throughout the video, as one of 36 locations.}
  \vspace{-1em}
  \label{fig:cater}
\end{figure}

We first evaluate our model on the CATER task.  CATER is composed of videos of CLEVR~\citep{johnson2017clevr} objects moving and, importantly, occluding each other.  Conceptually, it is meant to simulate the shell game, where the objective is to locate the final position of a ball that is hidden under a cup and shuffled with other cups.  In the CATER task, larger objects can occlude smaller objects visually, with respect to the camera viewpoint.  Furthermore, objects can also completely cover up smaller objects and move around with them, effectively moving multiple objects at once while occluding them entirely from view.  As in the shell game, the CATER task is to locate the final position of a unique golden object, named the ``snitch", where locations are represented as units on a 6x6 grid.  Therefore, the problem can be formulated as a 36-way classification problem. Whereas the CATER dataset features an additional split where the camera viewpoint can also move freely over the course of the video, we focus on the static camera setting in our experiments.  A visual demonstration of a sample CATER task is provided in Figure \ref{fig:cater}.

Each CATER video contains 301 frames.  We follow the experimental setup in ALOE~\citep{ding2021attention} and randomly sample 80 frames, which are then sorted to preserve temporal order. Each frame is resized to 64x64 resolution.  We also perform evaluation on 80 frame subsequences, but space their indices as evenly as possible.  We utilize a batch size of 256, and find convergence in 7,000 steps.  We find that using 3500 warmup steps, a learning rate of $3e^{-4}$ with a linear scheduler, and the Adam optimizer~\citep{kingma2014adam} with weight decay factor 0.05, generally produces the best results.  For other hyperparameter settings, we either match the configurations used by ALOE, or select them from validation set performance.  When adapting our framework on CATER, we concatenate only the first token from the encoder output of each frame and expose it to the decoder.

\subsection{ACRE Reasoning Setup} \label{sec:acre}

We also evaluate our model on the ACRE task, which tests causal inference capabilities.  ACRE is based off the Blicket experiment from developmental psychology~\citep{gopnik2000detecting}.  In the original formulation, a machine lights up when certain objects, called ``Blickets" are placed on it.  Preschool-aged children are tasked with determining which objects are Blickets, given demonstrations of certain object configurations and their resulting effects on the machine.  In the ACRE task, CLEVR objects are placed on a fixed platform, which glows or remains dim depending on the ``Blicketness" properties of the objects.  The model is provided with six context frames demonstrating different combinations of objects and their corresponding platform state, as well as a query frame containing an object combination that the model is expected to predict the result of.

There are four types of reasoning tasks the model is tested on, categorized by question types: direct, indirect, screened-off, and backward-blocking.  The potential answers for each of these questions is either that the platform's state is on, off, or unable to be determined.  In a direct question, the query combination is previously observed during one of the provided context trials; the model must be able to recognize it from the context and reference it to retrieve the answer.  In an indirect question, the query combination is novel, and the result must be deduced from several frames.  In screen-off questions, the model must learn the dynamic that as long as one Blicket object is present, the entire combination would turn the machine on.  Lastly, in the backward-blocking questions, the model must correctly deduce that the query combination cannot be ascertained from the available contexts.

\begin{figure}[t]
  \centering
  \includegraphics[width=0.99\linewidth]{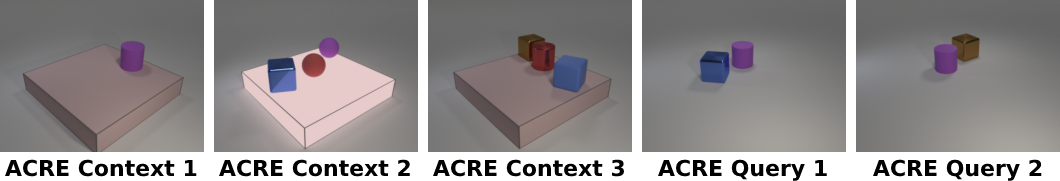}
  \caption{An example of the ACRE task.  The first three images represent context panels, which expose the state of the underlying platform to the model.  The model must predict the platform state for the two query configurations, shown as Query 1 and Query 2.  The result of Query 1 is \textit{undetermined}; although we know the purple cylinder does not light up the platform, we do not have enough information regarding the blue metal cube. For Query 2, the platform would not light up, which can be deduced from Contexts 1 and 3.  For ACRE, six context images are provided.}
  \vspace{-1em}
  \label{fig:acre}
\end{figure}

In our ACRE experiments, we reuse many of the same hyperparameter configurations from the CATER setup, except with 3000 warmup steps and 50,000 total training steps.

\subsection{Model Backbone}

Our proposed framework is general; beyond unifying recognition and reasoning tasks under a common framework, it enables us to evaluate the effect of different design decisions on higher-level reasoning performance in a principled manner.  We first investigate the effect of different encoder backbones used to process the visual input.  Specifically, we compare between two approaches: a ResNet Transformer (ResNet-T), where input images are processed as intermediate features of a ResNet-C model before being passed into a Transformer for further encoding, and a Vision Transformer (ViT), where input images are encoded as linear projections of pixel patches before being passed into a ViT-B model.

\begin{table}[ht]
\centering
\resizebox{\columnwidth}{!}{%
\begin{tabular}{cccc}
\multicolumn{1}{c}{\bf Encoder}  &\multicolumn{1}{c}{\bf Recognition Task}   &\multicolumn{1}{c}{\bf \# tokens}    &\multicolumn{1}{c}{\bf CATER}
\\ \hline 
ResNet-T & None (Ground-Up)& 1     & 18.10\%\\
ViT & None (Ground-Up)& 1     & 4.36\%\\
ResNet-T & LA-CATER ObjDet   & 1 & \textbf{74.09\%}\\
ViT & LA-CATER ObjDet   & 1  & 56.64\%\\
ResNet-T & LA-CATER ObjDet   & 10 & 71.68\%\\
ResNet-T & MS-COCO ObjDet & 1 & 69.79\%\\
\end{tabular}%
}
\vspace{-.5em}
\caption{Ablations on the CATER benchmark (static camera).  We demonstrate that choices between encoder architectures 
are considerable; using a ResNet-T encoder backbone consistently achieves better reasoning performance.  We ablate over the recognition task and between not using one, training on LA-CATER object detection, as well as MS-COCO object detection.  We also demonstrate that for the CATER task, performance does not improve when using more than 1 bottleneck token.  This is encouraging, as it shows that the information for solving a complex spatiotemporal reasoning task can be condensed into a small number of tokens.} 
\label{cater_ablations}
\end{table}

To better illuminate how the model backbone might affect visual recognition and reasoning separately, we first evaluate these two model backbones with respect to a visual recognition task such as object detection.  Because we then plan on evaluating our model on a reasoning task such as CATER, we perform object detection and report its results on frames that are visually similar to CATER scenes. As the ground truth bounding box annotations are not present in the CATER dataset, we use the LA-CATER dataset~\citep{shamsian2020learning}, which is generated to be identical to the CATER domain. It features the same camera configuration, the same usage of CLEVR objects, and the same number of possible objects per scene.


When pretraining on individual LA-CATER frames for the object detection task, we ask the model to predict all objects, including the ones that are occluded or contained. The task is thus inherently under-specified. We find that both architectures perform similarly; despite the occlusions, the ResNet Transformer achieves 81.53 AP50 whereas the Vision Transformer achieves 82.44 AP50. Wowever, we uncover a surprising result in Table~\ref{cater_ablations} that the ResNet Transformer encoder backbone consistently achieves better reasoning performance compared to the Vision Transformer encoder backbone.  We therefore believe that whereas the distinction between processing visual input using convolution and linear projection of patches may not be obvious for processing visual input, a convolutional inductive bias may still be important for enabling reasoning.  
We believe this is a signal that architectural considerations will be important when designing solutions to reasoning problems, and that simply reusing settings that work on pure visual recognition tasks can potentially inhibit performance.

\subsection{Number of Tokens}

\begin{figure}[h]
  \centering
  \includegraphics[width=0.99\linewidth]{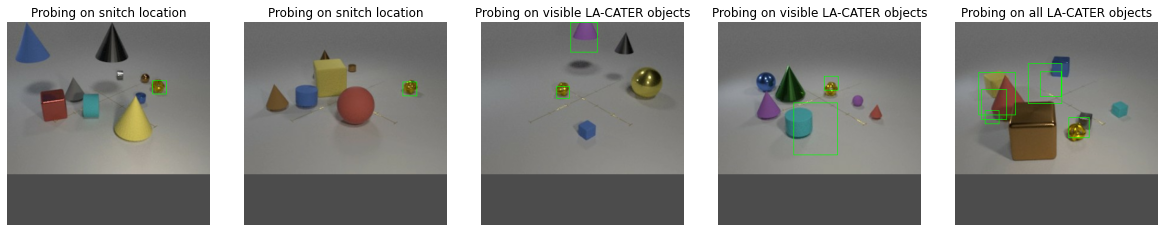}
  \caption{Visualizing the object-centric information in a frozen, learned CATER embedding via probing.  We demonstrate the result of predicting the snitch location, as well as the locations of other objects in the scene.  We find that the frozen CATER embedding is adept at modeling the location of the snitch, which is necessary and sufficient to solve the reasoning task of interest. We observe that the single embedding does not seem to accurately encode the locations of the other objects not relevant to the reasoning task.}
  \label{fig:snitch}
\end{figure}

Furthermore, we explore the other relevant inductive biases that could affect the final reasoning performance in our framework, such as the number of tokens produced per frame by the encoder.  We discover that not only does using a singular token work well for the CATER task, it actually outperforms using multiple tokens, as reported in Table~\ref{cater_ablations}.  This is potentially due to the singular token containing enough information to solve the task of locating the snitch.  To provide intuition on this, we visualize if the token contains an accurate belief over the location of the snitch at arbitrary frames in our probing experiments below, in Figure~\ref{fig:snitch}.  We visually confirm that the embedding does encode the snitch location, which is necessary and sufficient to solve the reasoning task of interest.  However, when probing on all visible objects, or all objects in a particular scene, we observe that the single embedding does not seem to accurately encode the locations of other objects. 
 The details of how probing is performed is elaborated upon in Section~\ref{probing}.

\subsection{Visual Recognition Objectives}
\label{sec:pretraining}

We aim to understand how the visual recognition pretraining objectives affect the reasoning capability.  As a default baseline, we first attempt to train our proposed framework to solve the CATER task directly.  In this baseline, no recognition task information is provided anywhere to the framework; it is tasked with trying to solve the reasoning task directly from the provided dataset with random weight initialization.  We refer to this baseline approach as “Ground-Up”.  In Table~\ref{cater_ablations}, we discover that it struggles to perform reasoning; our model achieves 18.10\% test accuracy when using a ResNet Transformer encoder backbone, and a mere 4.36\% when using a pure Vision Transformer encoding backbone.  For reference, a random guess on the location would be accurate 2.78\% of the time. 

We then evaluate our multi-task model formulation, by training on the object-detection task before learning how to perform reasoning.
Utilizing LA-CATER enables us to first train our framework on the recognition task of object-detection in a visually consistent setting, before proceeding to learn the desired reasoning task.  We discover that our end-to-end model trained in this way is able to perform the CATER task with 74.09\% test accuracy using a ResNet-T encoder. Our end-to-end multitask model is therefore able to outperform the default ALOE model, which achieves 70.6\% test accuracy.  Its performance is also comparable with the ALOE model that uses a task-specific L1 loss, which achieves $74.0 \pm 0.3\%$ (See Appendix). 
In contrast, we do not use any extra auxiliary task-specific losses in our setup.  We therefore conclude that the addition of the general object-detection task helps the model perform spatiotemporal reasoning in an end-to-end unified model.  Given ALOE's finding that object-centric abstractions of the input are important for reasoning, which they leverage a pre-trained MONET model to generate, we take this as a preliminary sign that the model forms strong object-centric representations implicitly. 

\begin{table}[h]
\centering
\resizebox{\columnwidth}{!}{%
\begin{tabular}{cccc}
\multicolumn{1}{c}{\bf Dataset} &\multicolumn{1}{c}{\bf Recognition Task}  &\multicolumn{1}{c}{\bf ACRE}   &\multicolumn{1}{c}{\bf CATER}
\\ \hline
LA-CATER & Detect All Objects  & 67.27\%   & \textbf{74.09\%}\\
LA-CATER & Detect Visible Objects & 81.65\%   & 73.44\% \\
LA-CATER & Count All Objects  & 44.04\%   & 68.36\%\\
LA-CATER & Count Unique Objects  & 41.48\%   & 64.78\%\\
LA-CATER & Snitch Detection  & 79.70\%   & 72.66\%\\
MS-COCO & Object Detection  & \textbf{83.81\%}   & 69.79\%\\
\end{tabular}%
}
\vspace{-.5em}
\caption{Comparisons across visual recognition objectives and datasets. We report results on the ``compositional'' split of ACRE and static camera split of CATER. We choose the objectives of Detect Objects to determine if recognition tasks that require strong spatial localization ability indeed help reasoning performance under our framework.  We contrast this property with other visual recognition objectives such as Count All Objects and Count Unique Objects in a scene; such perception objectives do need to understand the visual scene to some degree but do not explicitly require spatial localization to understand where in the picture each object is.  We also find a balance in the Snitch Detection task, which requires moderate spatial localization.  We generally find that utilizing visual recognition tasks that require understanding spatial localization result in the best reasoning performance.}
\label{objective_ablations}
\end{table}

We also investigate how different types of visual recognition tasks can enable reasoning capabilities for ACRE in Table~\ref{objective_ablations}.  We therefore ablate over a variety of different tasks.  In LA-CATER Object Detection, the visual recognition is tasked with predicting the bounding boxes and shapes of all objects in the scene, hidden or visible; for MS-COCO the model predicts the bounding boxes and class labels.  In Detect Visible Objects, the model is only asked to predict the bounding boxes and shape labels for all visible objects from the camera perspective.  In Count All Objects and Count Unique Objects, the model is tasked with predicting the number of total objects in the scene, and the number of unique shapes in the scene, respectively; notably, these objectives do not require the model to learn strong spatial localization.  Lastly, the Snitch Detection objective encourages the model to predict the bounding box location of the golden snitch.  We find that the best performance on ACRE is achieved through visual objectives that involve spatial localization.  Indeed, the top two performances on both ACRE and CATER utilize the Object Detection task: on MS-COCO and visible objects from LA-CATER for ACRE, and LA-CATER and visible LA-CATER for CATER.  Learning to predict hidden or occluded objects, as in the complete LA-CATER Object Detection, may not be helpful for the ACRE task since there are no fully occluded objects in the environment. We also notice that the visual recognition tasks that do not require explicit spatial localization, such as Count All Objects or Count Unique Objects, perform worst on their respective reasoning environments.  Furthermore, we find that the snitch detection performs well; on CATER, this is to be expected since it is directly useful for the final reasoning objective.  Overall, these results support the hypothesis that recognition tasks that learn spatial localization of objects are beneficial for unlocking reasoning capabilities.

\begin{figure*}
  \centering
  \includegraphics[width=0.9\linewidth]{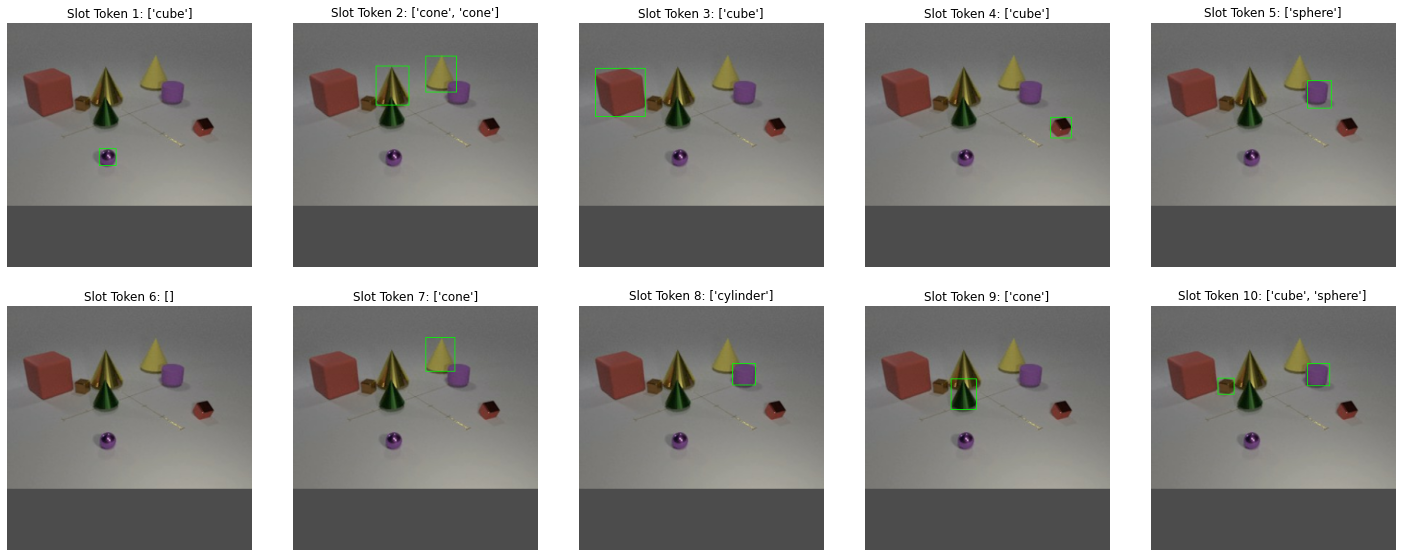}
  \caption{We visualize the true positive object and shape predictions for the ten token embeddings in a frozen ResNet-T encoder using probing. Each token appears to be adept at encoding one or a few objects well, and different tokens tend to focus on different objects.}
  \label{fig:probes}
\end{figure*}


\subsection{Pretraining Dataset}
We further explore whether object-detection as an objective is generally useful, or if the benefits are only limited to the visual domain of the reasoning task.  In other words, we investigate if there are inherent properties of object detection that could potentially enable the learning of object-centric abstractions that can extend beyond visual domains, and still be useful in enabling reasoning capabilities.  We test this in our model by evaluating CATER performance with object-detection pre-training on the MS-COCO dataset.  Composed of natural images, the MS-COCO dataset is visually dissimilar from the synthetic CATER environment, and the two datasets share no overlapping object classes.  However, as reported in Table~\ref{objective_ablations}, we find that our method still achieves 69.79\% on CATER test performance, and 83.81\% ACRE test performance.  Therefore, although it is trained for object detection on an entirely different visual domain, our method is still able to exhibit reasoning capabilities, and even beats the default ALOE performance.  Whereas ALOE requires pretraining a separate model on the same visual demonstrations from the same domain in a pipeline-style approach, we show that our model is still able to perform reasoning in an end-to-end way, leveraging object-detection capabilities on out-of-domain data. Our results hint at the efficacy of object detection as a useful general task.

\subsection{Probing}
\label{probing}

We perform probing to understand the information encoded in the learned embeddings.  We begin by investigating the singular bottleneck embedding learned by a CATER model under our proposed framework that is pretrained with an LA-CATER Object Detection task and utilizes a ResNet-T backbone.  We load and freeze the encoder of the learned CATER model, and proceed to train an autoregressive decoder on the snitch detection task, LA-CATER Visible Object Detection, and LA-CATER Object Detection.  We then visualize our predictions in Figure \ref{fig:snitch} for randomly selected frames, and demonstrate qualitatively that the learned embedding consistently encodes an accurate prediction of where the snitch is at any given time, regardless of the probing objective.  What this shows is that the frozen bottleneck always encodes the snitch location information, for which a learned probing autoregressive decoder can be cheaply trained to read out.  As tracking where the snitch is over multiple frames is indeed a necessity to solve the CATER task, and it is therefore a nice result that such information can be shown to be stored in the singular bottlenecking token.


We then perform probing experiments on a ResNet-T encoder that was previously trained on LA-CATER for the object detection task using ten bottleneck embeddings to understand how the reasoning models for CATER and ACRE are initialized.  As we utilize this object detection pretraining procedure before each of the two reasoning tasks we evaluate on, the embedding outputs of this encoder are especially interesting to investigate in order to understand how they can potentially enable reasoning capabilities.  We freeze the encoder, and proceed to learn an autoregressive decoder on the LA-CATER object detection task, but selecting one token at random each iteration.  After convergence, we use the trained decoder to evaluate each individual bottleneck token of the encoder by qualitatively visualizing their predicted bounding boxes on a random LA-CATER test set frame.

In Figure \ref{fig:probes_annotated}, in Appendix~\ref{app:probing}, we display all the predicted bounding boxes with a confidence score above 90\% for each token.  We notice three things - firstly, no token is encoding every object. 
In fact, there exists a token, token number 6, that does not appear to encode any object strongly.  This is not of particular issue, as there are ten total ``slot" tokens learned, but only eight objects in the scene.
Secondly, we discover that with the exception of number 6, the tokens are able to predict one or a few objects exceptionally accurately.  We visualize these clean predictions in Figure \ref{fig:probes}, and display their corresponding predicted shapes as well.  These results demonstrate that individual tokens are able to not only understanding where certain objects are but their properties as well. 
Lastly, we observe that despite each token representing one or a few objects well, all of the objects in the scene are accounted for and there are few overlaps.  Therefore we discover 
that in combination, the tokens learned under our framework are able to collectively represent an entire scene in terms of its objects well, while encoding object-centric information individually.
\section{Conclusion and Future Work}

We attempt to build a unified framework for visual recognition and reasoning with general-purpose Transformers. We hypothesize that object detection motivates the network to learn object-centric representations which are beneficial for visual reasoning. We leverage a transformer-based sequence prediction framework to jointly tackle detection and reasoning on two diagnostic datasets, CATER and ACRE. Our quantitative and qualitative results show encouraging signs that object detection indeed helps visual reasoning, and that object-centric representations seem to emerge from object detection pretraining. Interestingly, our experiments also reveal that although different inductive biases
may have little impact on object detection performance, their relative gaps on the reasoning benchmarks are nontrivial. We hope that our findings can bring awareness and consideration to reasoning performance when designing network architectures, as well as motivate further explorations on building unified multi-task models for perception and reasoning. In the future, we would like to explore other recognition tasks and joint training strategies.

\noindent\textbf{Limitations:} Our probing experiment is qualitative. Quantitative metrics are needed to measure the ``locality'' and ``completeness'' of detected objects. The diagnostic datasets are synthetic and moderate-scale, it is unclear if a similar trend holds with large-scale pre-training on natural images.

\noindent\textbf{Acknowledgements:} Part of the research was conducted while Calvin Luo worked as a student researcher at Google. Calvin Luo and Chen Sun are in part supported by research grants from Honda Research Institute, Meta AI, Samsung Advanced Institute of Technology, and a Richard B. Salomon Faculty Research Award.

\bibliography{ref.bib}
\bibliographystyle{latex/icml2023}

\newpage
\appendix
\onecolumn
\section{Appendix}
\subsection{Comparison with state-of-the-art}
Below, we list the benchmark results on ACRE (Table \ref{acre_benchmark}) and CATER (Table \ref{cater_benchmark}).  Our proposed model is competitive with the state-of-the-art on CATER, as it underperforms compared to OPNet~\citep{shamsian2020learning} by .7\%. However, this difference can be justified by considering the task-engineered architectural design decisions behind OPNet. The OPNet model is composed of a perception module and two reasoning modules tailored specifically for the CATER task; one reasoning module determines what object to track, and another handles the target in the case of full occlusion and determines where its location is. This architecture would not translate well to other reasoning tasks, such as ACRE for example, where tracking and/or occlusions are not relevant. We can therefore, by design, expect OPNet to have a strong performance on CATER - and the fact that our model achieves competitive performance with the state-of-the-art while making no task-specific design decisions whatsoever is encouraging. Indeed, our model can extend to ACRE as well, with no architectural adjustment.

On the other hand, ALOE~\citep{ding2021attention} is similar to our approach in that the reasoning module is modeled as a general transformer architecture that can be applied to multiple reasoning tasks without incorporating inductive biases. However, ALOE still relies on a separately trained perceptual model, and still resembles a two-stage pipeline. We improve on this by proposing a singular, unified architecture that learns strong implicit object-centric representations (as demonstrated by probing), while also being able to solve reasoning tasks. We find that we are able to match ALOE’s performance, empirically, while avoiding using any task-specific losses (like the L1 loss).

Our model is not hand-designed for a particular task, unlike OPNet, and does not require any separately trained perceptual models, unlike ALOE - given these improvements over the existing methods, we find it encouraging that we can still achieve strongly competitive performance in comparison.

\begin{table}[ht]
\caption{Benchmark results on ACRE.  Numbers apart from our method are taken from the ALOE~\citep{ding2021attention} paper.  We show that our end-to-end unified architecture is able to achieve competitive results with the current state-of-the-art on causal reasoning.}
\label{acre_benchmark}
\begin{center}
\begin{tabular}{ll}
\multicolumn{1}{c}{\bf Model}  &\multicolumn{1}{c}{\bf ACRE (Comp)}
\\ \hline \\
CNN-BERT & 43.79\%\\
NS-OPT & 69.04\%\\
Aloe               & \textbf{91.76\%}\\
\hline
Our Method         & 83.81\%\\
\end{tabular}
\end{center}
\end{table}

\begin{table}[ht]
\caption{Benchmark results on CATER.  Numbers apart from our method are taken from the ALOE~\citep{ding2021attention} paper.  We show that our end-to-end unified architecture is able to achieve competitive results with the current state-of-the-art on a complex spatiotemporal reasoning task.}
\label{cater_benchmark}
\begin{center}
\begin{tabular}{ll}
\multicolumn{1}{c}{\bf Model}  &\multicolumn{1}{c}{\bf CATER Top 1 (Static)}
\\ \hline \\
R3D LSTM           & 60.2\%\\
R3D + NL LSTM      & 46.2\%\\
OPNet              & \textbf{74.8\%}\\
Hopper             & 73.2\%\\
Aloe (no auxiliary)& 60.5\%\\
Aloe               & 70.6\%\\
Aloe (with L1 loss)& $74.0 \pm 0.3$\%
\\ 
\hline
Our Method         & 74.1\%\\
\end{tabular}
\end{center}
\end{table}

\subsection{Probing}
\label{app:probing}

\subsubsection{False Positives}
\begin{figure}
  \centering
  \vspace{-0.5em}
  \includegraphics[width=0.9\linewidth]{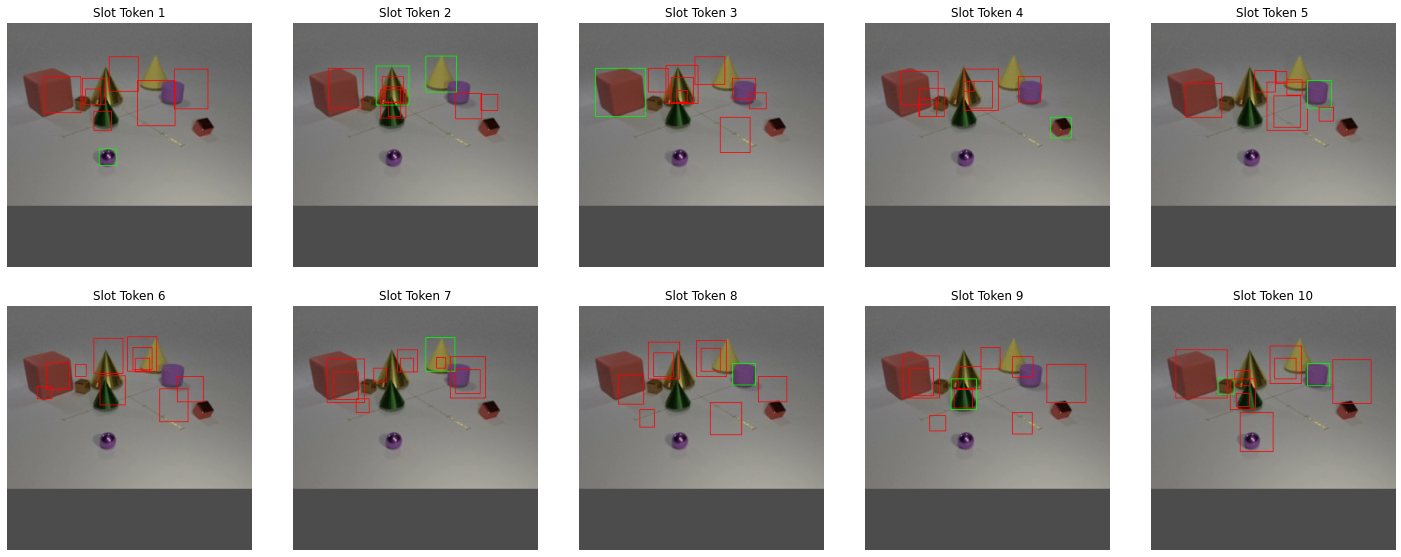}
  \caption{We visualize all bounding box predictions for each token in a frozen ResNet + Transformer encoder using probing.  We highlight the true positive bounding box predictions in green.}
  \vspace{-1em}
  \label{fig:probes_annotated}
\end{figure}

Probing is performed on the embeddings produced by the encoder of a trained model under our framework, to understand the information learned to be useful for the downstream task.  In particular, we perform probing under our framework by loading and freezing the encoder, but learning a separate decoder for object-detection on LA-CATER frames that conditions on a randomly selected token at each iteration.  Object detection is used to understand what objects are accurately localized in the embeddings.  Then, the decoder is used to visualize a desired LA-CATER frame of choice, for each embedding produced by the encoder.

Above, in Figure~\ref{fig:probes_annotated}, we display all bounding boxes with a confidence score above 90\% for each token.  These represent where each token believes there to be objects in the scene.  Highlighted in green are the bounding boxes that indeed match the ground truth, representing the objects that the embedding is accurately encoding.  These green bounding boxes are highlighted in Figure~\ref{fig:probes}.

The high rate of false positives can be attributed to a number of factors.  Firstly, we perform our probing on object-detection with hidden object annotations, rather than only visible objects.  This causes the model to predict false positive predictions of bounding boxes within other bounding boxes, even if it is not the case.  Furthermore, given the object detection task, it is not naturally disincentivized to only predict a few bounding boxes; given that the amount of objects in each frame of the training data is large, it is encouraged to predict multiple objects even if it does not encode the relevant information accurately, resulting in false positive predictions. 

\subsubsection{Additional Visualizations}
\begin{figure}
  \centering
  \includegraphics[width=0.95\linewidth]{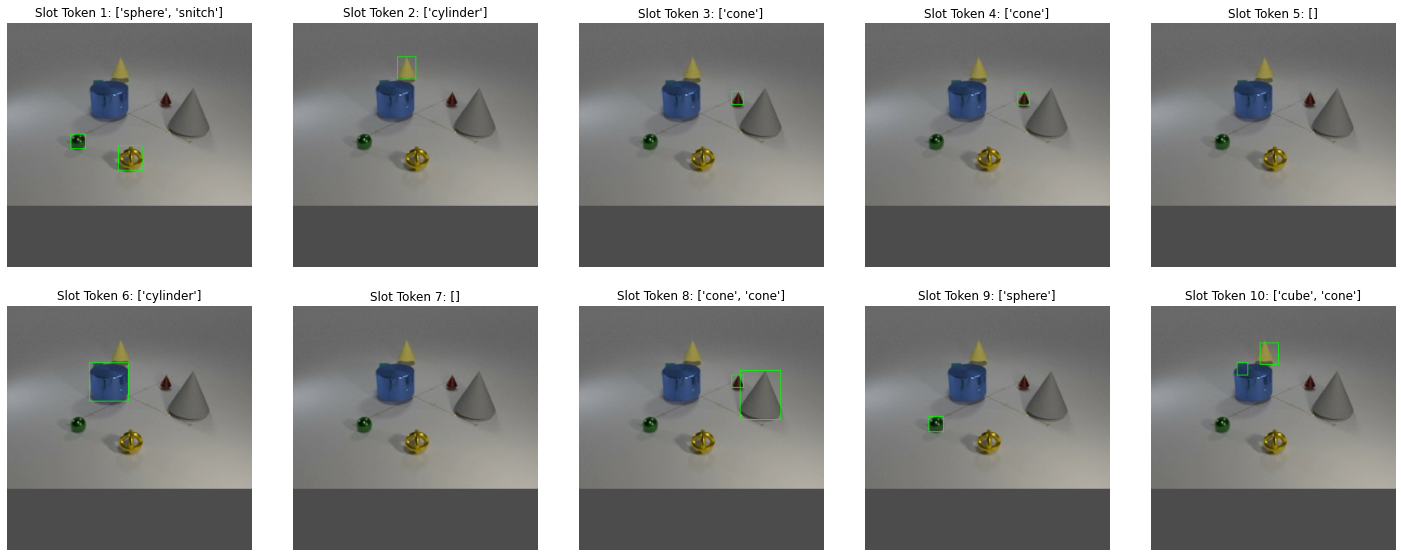}
  \includegraphics[width=0.95\linewidth]{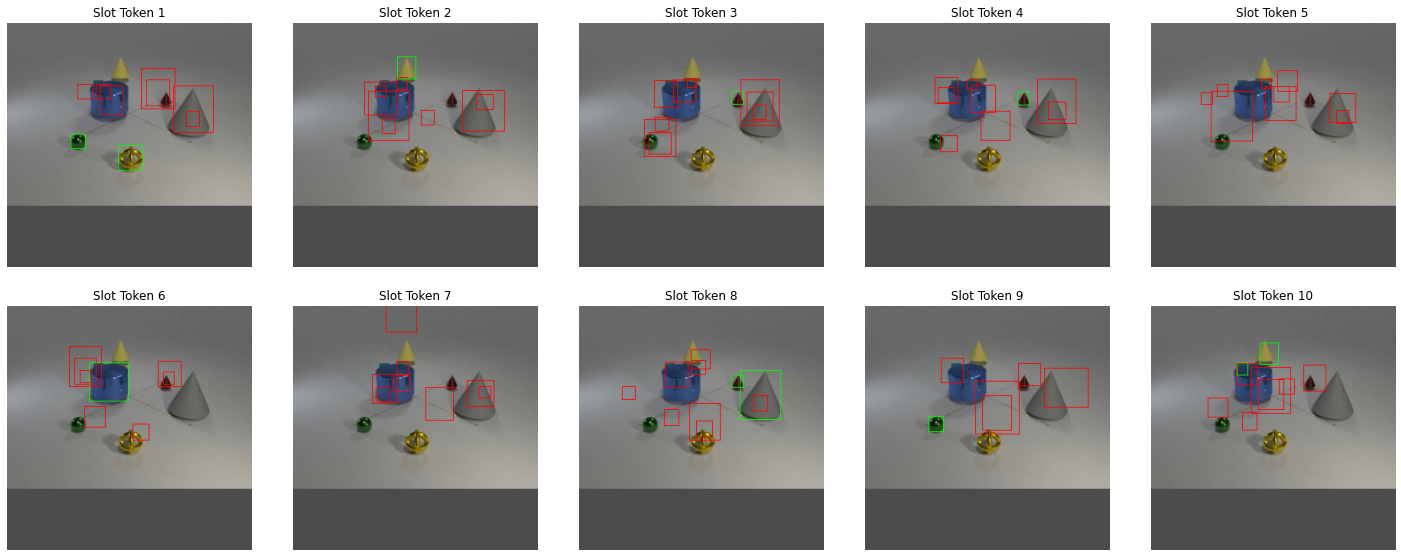}
  \caption{Sample test frame number 4 for probing from the LA-CATER dataset.}
  \label{fig:test_4_probes}
\end{figure}

We further include bounding box and shape predictions for each of the ten slot embeddings in a frozen ResNet + Transformer encoder on 6 randomly sampled test frames from the LA-CATER dataset.  These figures visualize what object-centric information each individual frozen slot embedding is encoding.  As mentioned before, we notice each of the embeddings encode at most a few objects, and all of the objects in a scene are encoded by at least one embedding.  Furthermore, we observe that the model is even able to recognize heavily occluded objects such as the small blue cube hidden behind the large blue cylinder in Figure \ref{fig:test_4_probes}, which is successfully captured by Slot Token 10.

\begin{figure}[ht]
  \centering
  \includegraphics[width=0.95\linewidth]{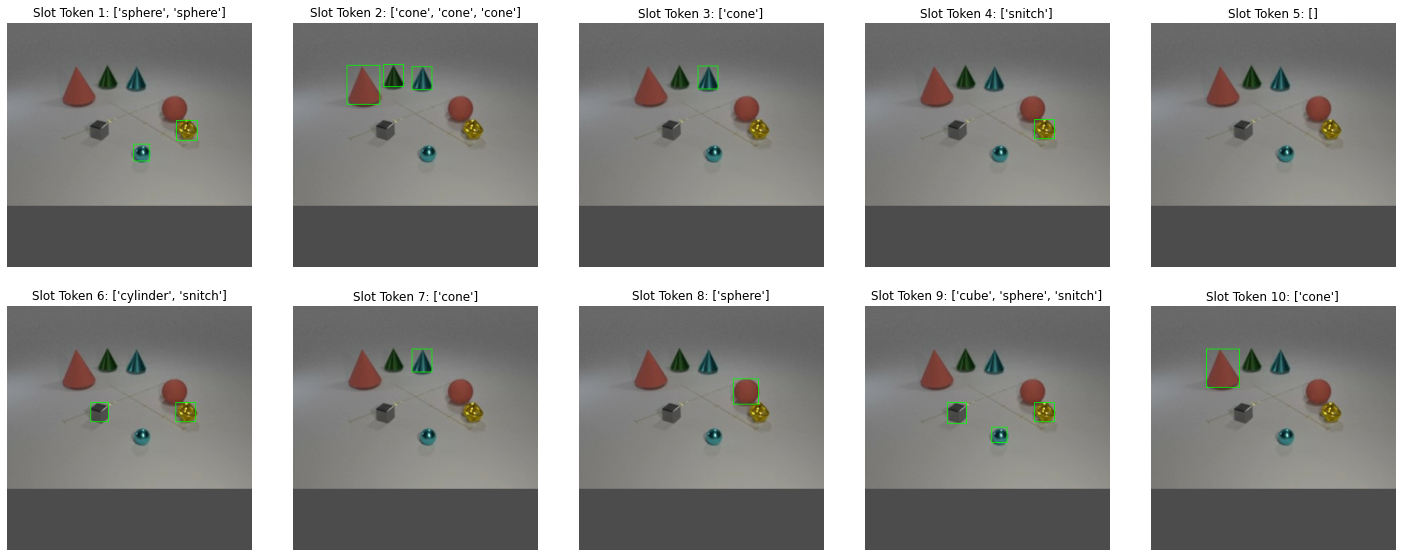}
  \includegraphics[width=0.95\linewidth]{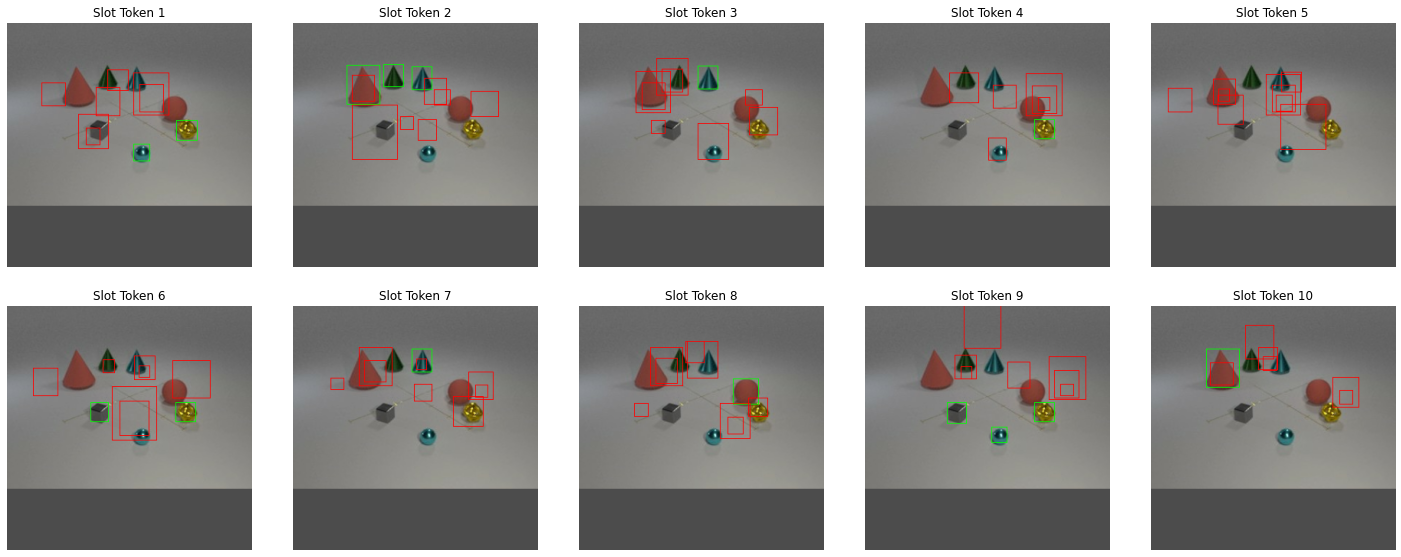}
  \caption{Sample test frame number 1 for probing from the LA-CATER dataset.}
  \label{fig:test_1_probes}
\end{figure}
\begin{figure}
  \centering
  \includegraphics[width=0.95\linewidth]{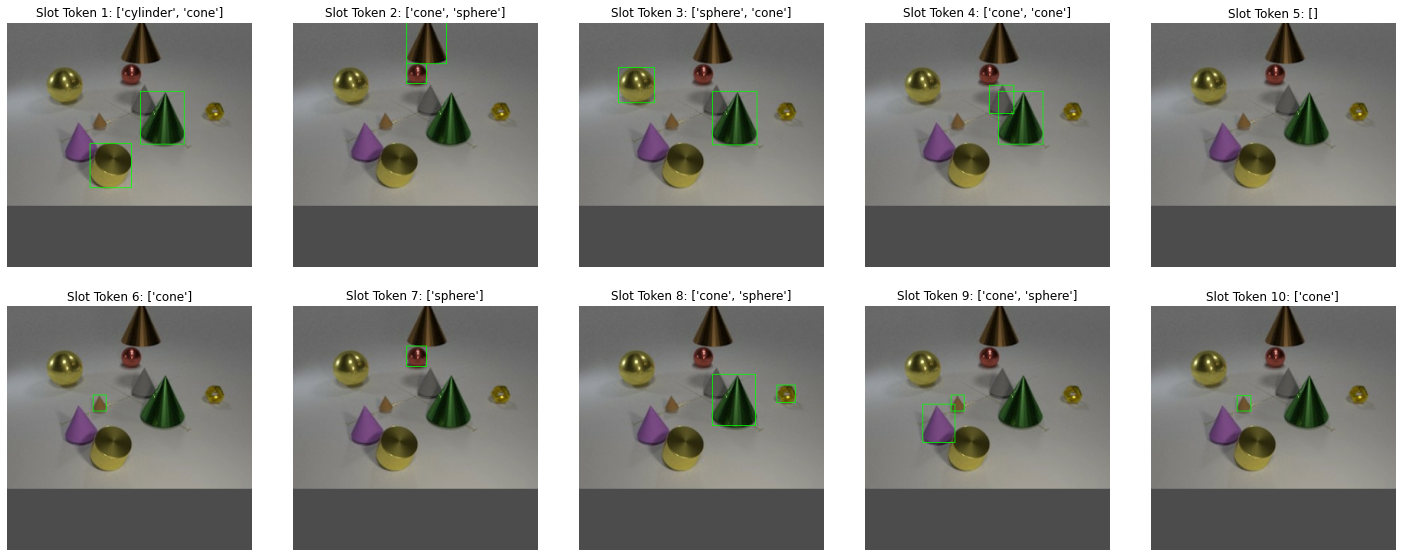}
  \includegraphics[width=0.95\linewidth]{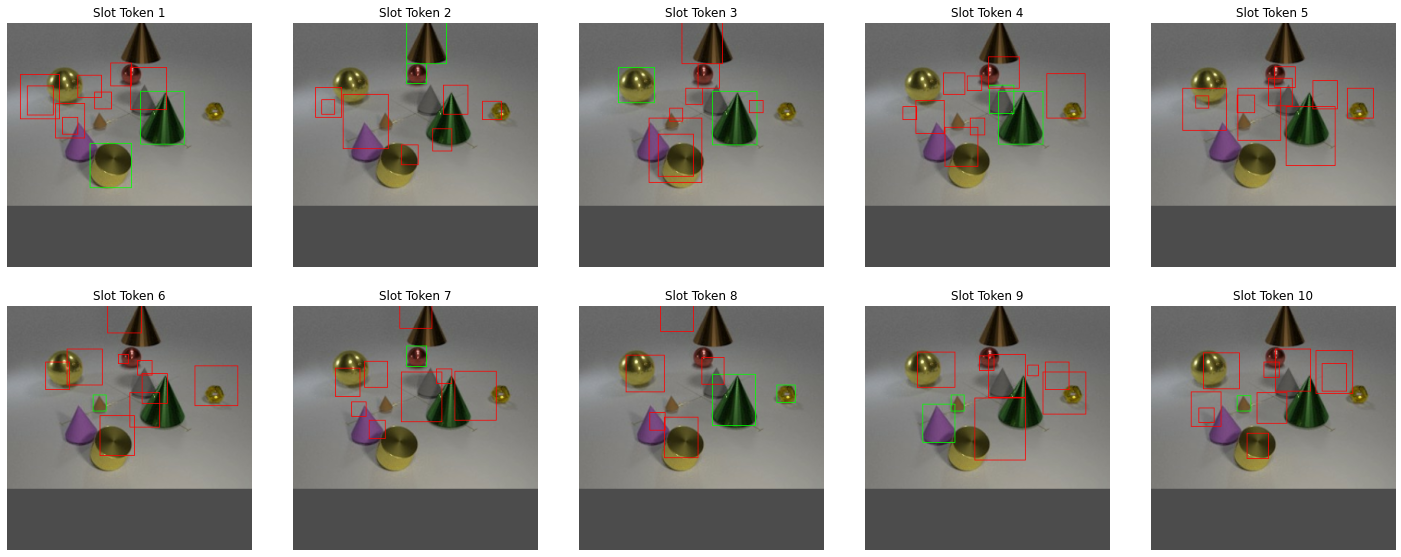}
  \caption{Sample test frame number 2 for probing from the LA-CATER dataset.}
  \label{fig:test_2_probes}
\end{figure}
\begin{figure}
  \centering
  \includegraphics[width=0.95\linewidth]{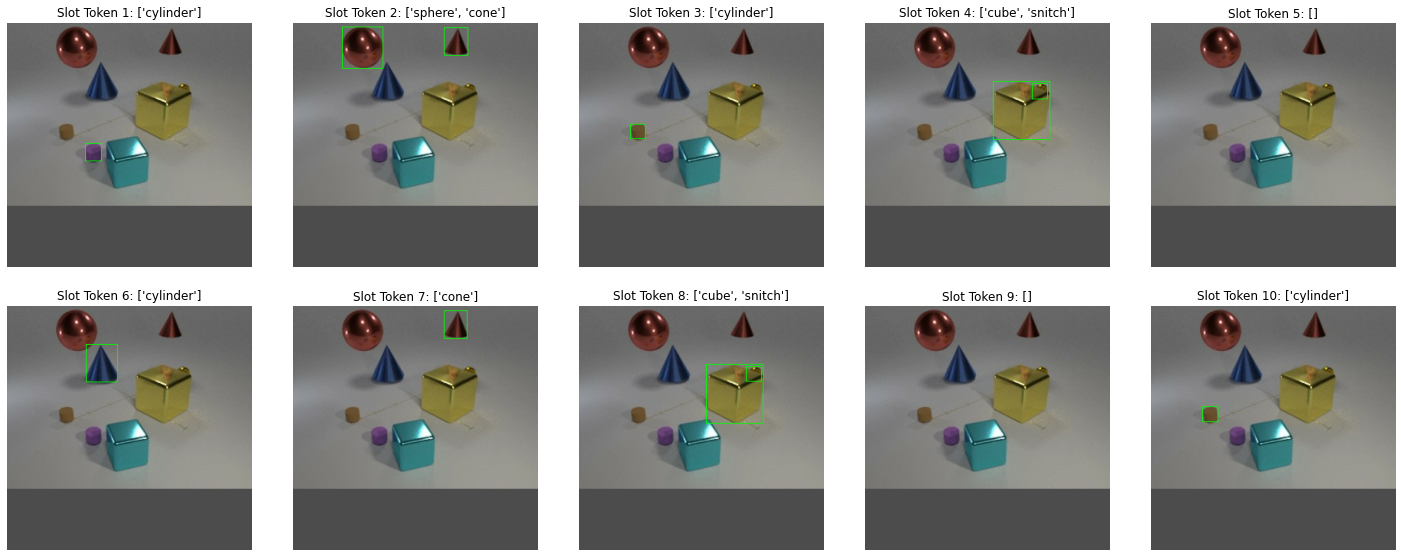}
  \includegraphics[width=0.95\linewidth]{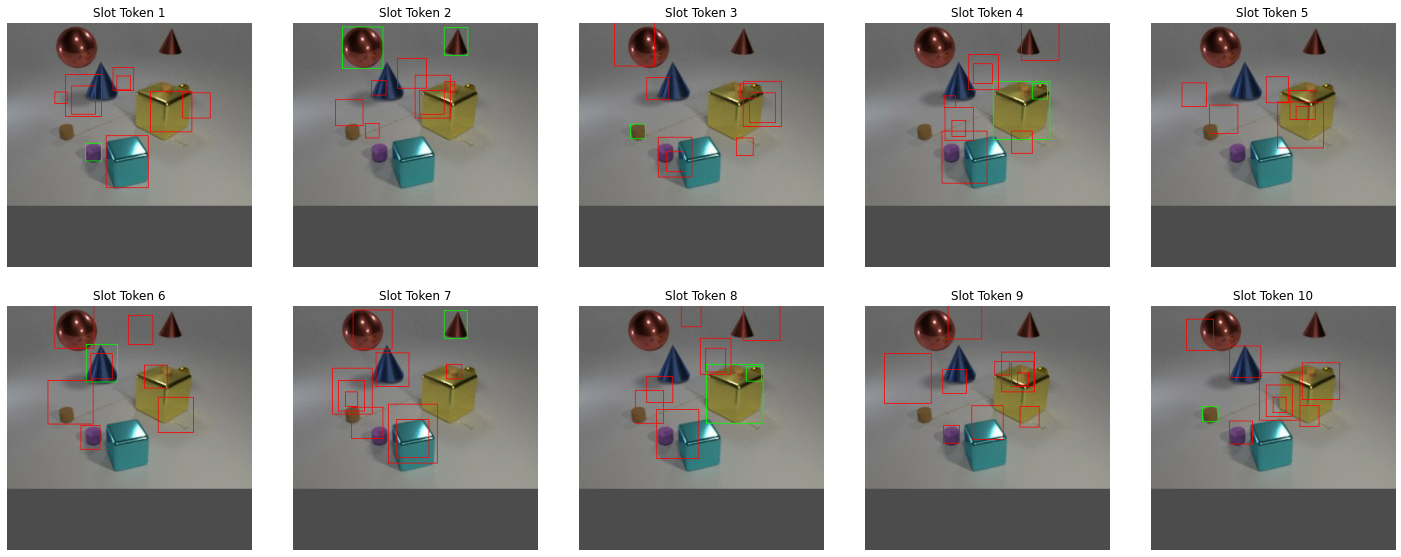}
  \caption{Sample test frame number 3 for probing from the LA-CATER dataset.}
  \label{fig:test_3_probes}
\end{figure}
\begin{figure}
  \centering
  \includegraphics[width=0.95\linewidth]{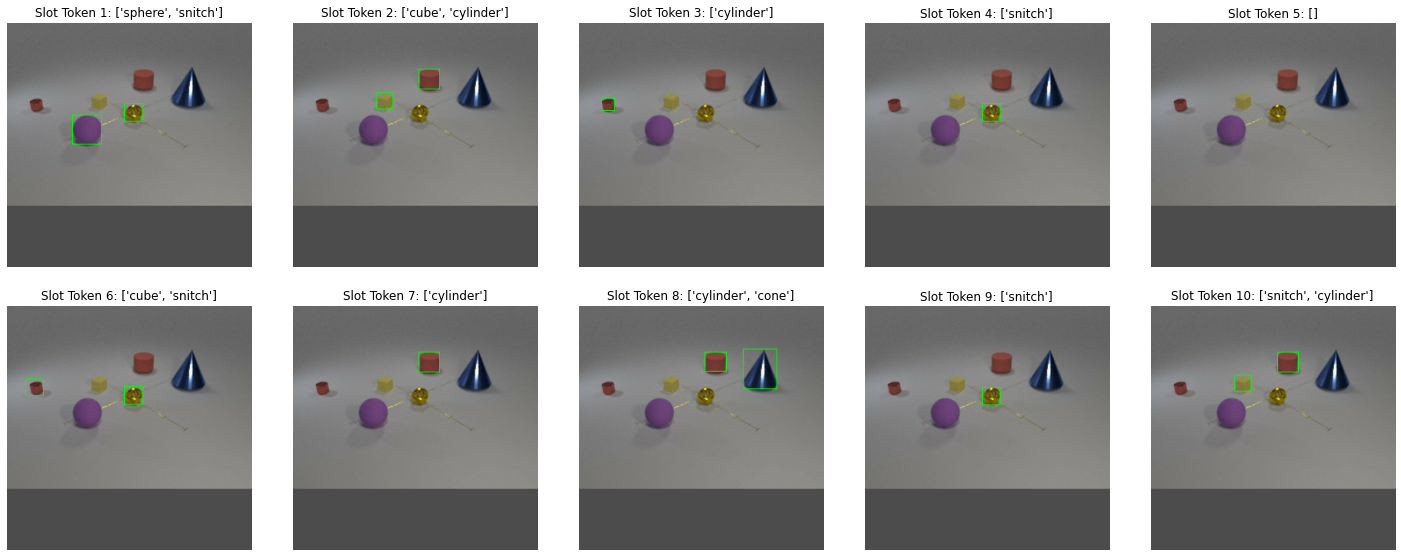}
  \includegraphics[width=0.95\linewidth]{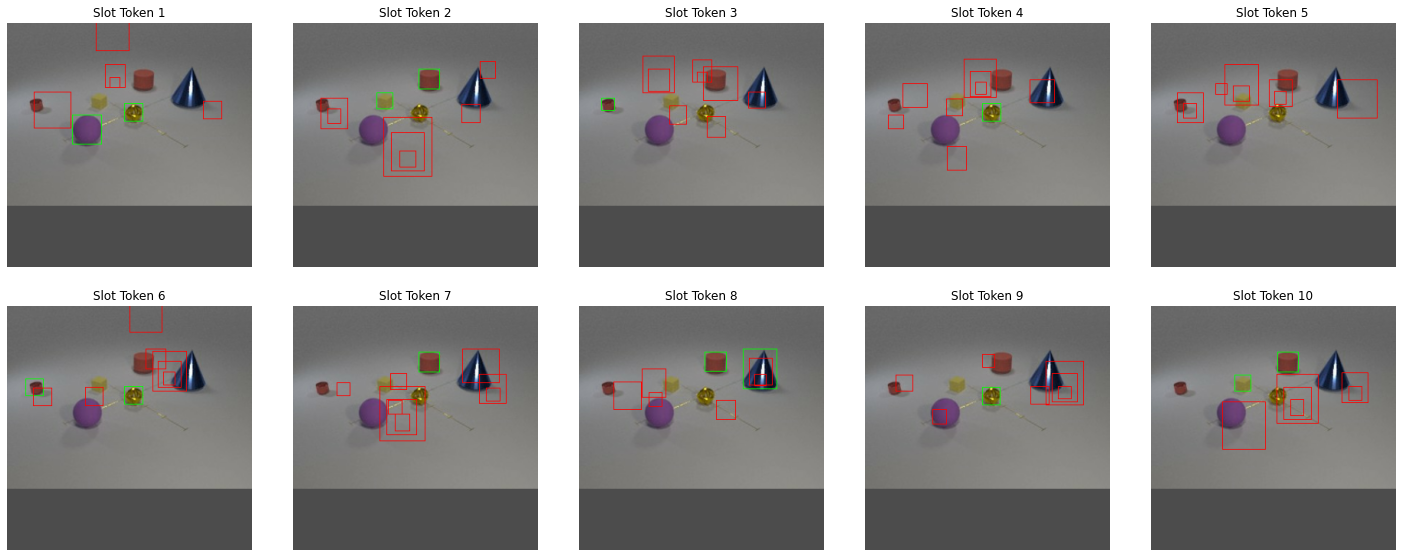}
  \caption{Sample test frame number 5 for probing from the LA-CATER dataset.}
  \label{fig:test_5_probes}
\end{figure}
\begin{figure}
  \centering
  \includegraphics[width=0.95\linewidth]{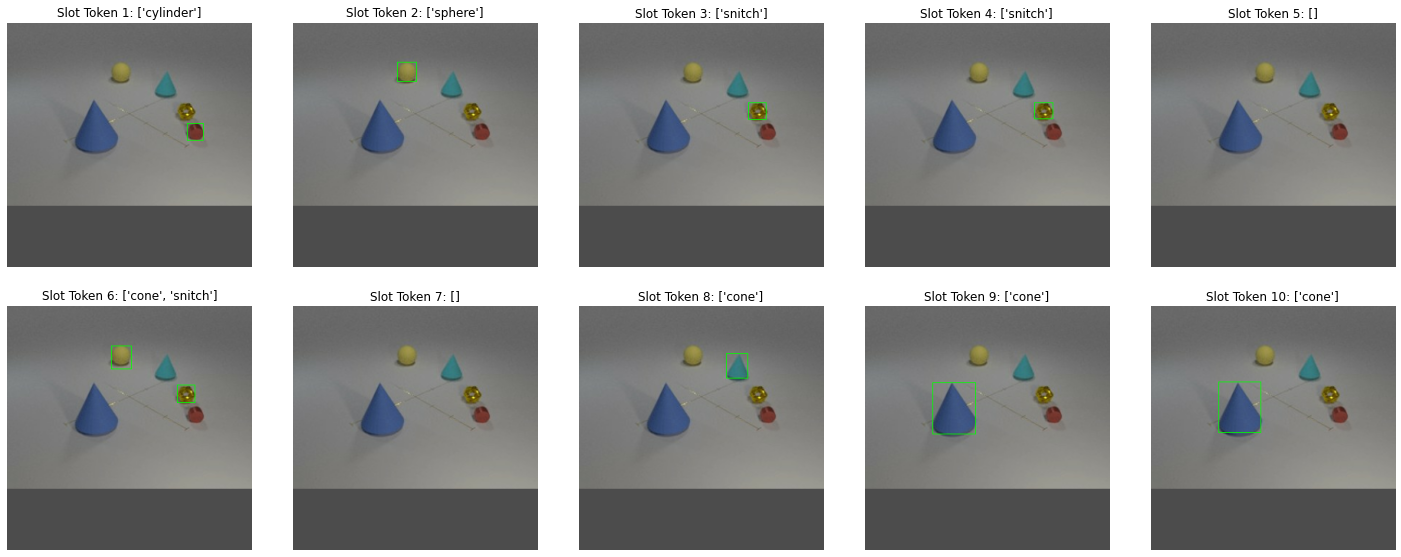}
  \includegraphics[width=0.95\linewidth]{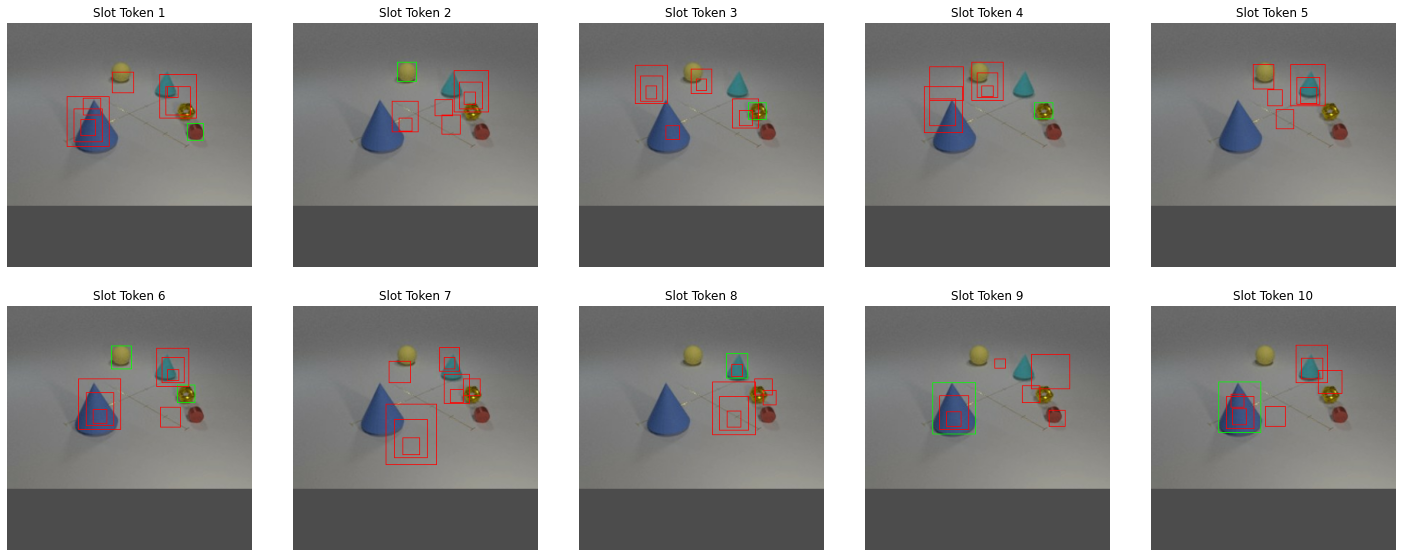}
  \caption{Sample test frame number 6 for probing from the LA-CATER dataset.}
  \label{fig:test_6_probes}
\end{figure}

\end{document}